\documentclass{article}

\usepackage{microtype}
\usepackage{graphicx}
\usepackage{booktabs} % for professional tables
\usepackage{cuted}

\usepackage{hyperref}
\usepackage{subcaption}
\usepackage{amsfonts}
\usepackage{amsmath}
\usepackage{xfrac}
\usepackage[super]{nth}

\newcommand{\fea}[1]{\boldsymbol{#1}}
\newcommand{\sfea}[2]{\boldsymbol{#1_{#2}}}
\newcommand{\bfea}[1]{\overline{\boldsymbol{#1}}}

\usepackage[accepted]{icml2020}

\icmltitlerunning{PackIt: A Virtual Environment for Geometric Planning}

\begin{document}

\twocolumn[
\icmltitle{PackIt: A Virtual Environment for Geometric Planning}



\icmlsetsymbol{equal}{*}

\begin{icmlauthorlist}
\icmlauthor{Akit Goyal}{princeton}
\icmlauthor{Jia Deng}{princeton}
\end{icmlauthorlist}

\icmlaffiliation{princeton}{Department of Computer Science, Princeton University, Princeton, NJ, USA}

\icmlcorrespondingauthor{Ankit Goyal}{agoyal@princeton.edu}
\icmlcorrespondingauthor{Jia Deng}{jiadeng@cs.princeton.edu}

\icmlkeywords{Machine Learning, ICML}
\vskip 0.3in
]

\printAffiliationsAndNotice{}  % leave blank if no need to mention equal contribution

\begin{abstract}
The ability to jointly understand the geometry of objects and plan actions for manipulating them is crucial for intelligent agents. We refer to this ability as geometric planning. Recently, many interactive environments have been proposed to evaluate intelligent agents on various skills, however, none of them cater to the needs of geometric planning. We present PackIt, a virtual environment to evaluate and potentially learn the ability to do geometric planning, where an agent needs to take a sequence of actions to pack a set of objects into a box with limited space. We also construct a set of challenging packing tasks using an evolutionary algorithm. Further, we study various baselines for the task that include model-free learning-based and heuristic-based methods, as well as search-based optimization methods that assume access to the model of the environment. Code and data are available at \url{https://github.com/princeton-vl/PackIt}.
\end{abstract}

\section{Introduction}
A crucial component of human intelligence is the ability to \textit{simultaneously} reason about the geometry of objects and plan actions for manipulating them. This ability comes in handy in many everyday scenarios, like organizing utensils on a stand, putting stuff in boxes while moving out or rearranging objects in a room to bring in a new couch. In all these cases, we need to understand the geometry of objects and plan actions while taking care of various spatial constraints. We use the term \textit{geometric planning} to refer to this ability. As opposed to classical symbolic planning, which involves manipulating and reasoning about symbols, geometric planning involves manipulating and reasoning about geometric entities. It is desirable for artificial agents like robots to possess the ability of geometric planning so that they can operate in unconstrained human environments and assist us in day-to-day activities.

The first step for building artificial agents that possess a certain ability is to have a benchmark for evaluating that ability. Following this principle, several environments have been proposed in the AI literature which test agents on various abilities. For example 3D interactive environments like  AI2-THOR~\cite{ai2thor}, HoME~\cite{brodeur2017home} and CHALET~\cite{yan2018chalet} test agents on navigation and object interaction. Similarly, simulated environments in robotics~\cite{plappert2018multi} evaluate agents on low-level continuous control problems. However, none of the prior environments directly cater to geometric planning. In 3D interactive environments like AI2-THOR and HoME, the geometrical variability of the objects does not play a large role, and hence, an agent need not reason about the geometry of objects while interacting with them. Likewise, simulated environments in robotics do not require reasoning about the geometry of multiple real-world objects simultaneously as their focus is on low-level control problems like grasping.

To address the need for an environment that caters to geometric planning, we propose \textit{PackIt}. PackIt is a 3D virtual environment that explicitly tests an agent for geometric planning. Our hope is that the availability of such a benchmark would help the community to build and test components that would enable artificial agents to do geometric planning. Also, PackIt is OpenAI gym~\cite{openai} compatible, which makes it easily adoptable.

\begin{figure}[b]
  \centering
  \includegraphics[width=\columnwidth]{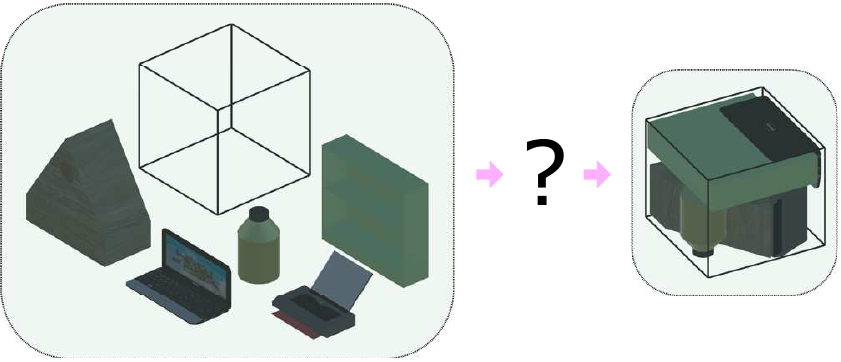}
  \caption{An example to demonstrate the packing task. Given a set of objects and a box, the task is to figure out a way of placing the objects completely inside the box.}
  \centering
  \label{fig:packing}
\end{figure}

Specifically, in PackIt, we choose the packing task as a proxy to evaluate geometric planning. The packing task requires an agent to perform a sequence of actions for placing a set of everyday objects (like tables, guitars, and bowls) in a given box (Fig. \ref{fig:packing}). The packing task serves as an excellent testbed for geometric planning for the following reasons. First, it requires understanding the geometric properties of real-world shapes like protrusions, contours, and holes. And second, it requires planning actions to manipulate real-world shapes while taking care of various spatial constraints.

The PackIt environment, unlike prior 3D interactive environments (like AI2-THOR, HoME, and CHALET), requires an agent to understand the geometry of objects while interacting with them. Also, unlike prior simulated environment in robotics, it requires an agent to reason about multiple real-world shapes simultaneously. To make the task tractable, we abstract away low-level motor actions, with high-level actions that assume perfect control. The advantage of this is that it allows us to focus on high-level geometrical planning skills without getting side-tracked by the difficulties in orthogonal problems like grasping.

To sum up, geometric planning in PackIt embodies the general ability to arrange shapes to satisfy constraints, which is required for complex high-level tasks such as grocery shopping, room decluttering and cleaning, and warehouse management. Another utility of PackIt is that it enables training learning-based methods for packing problems arising in ISO luggage packing~\cite{tiwari2010fast} and 3D printing~\cite{araujo2019analysis}. In these domains, just like PackIt, CAD models are directly fed to the system. These are scenarios a well-performing model in PackIt can generalize to.

In addition to the PackIt virtual environment, we also need challenging packing problems to evaluate an agent's skills. For this, we propose a novel way to automatically generate progressively harder packing problems by using an evolutionary algorithm. We generate a dataset of hard packing problems, which we refer to as the \textit{PackIt dataset}. 

To facilitate future research, we design model-free learning-based as well as heuristic-based baselines on the proposed task. In the model-free setup, the learning-based baseline outperforms the heuristic-based one, illustrating that learning can be a viable option to acquire geometric planning skills for packing. 

We also evaluate search-based optimization baselines by augmenting the model-free baselines with a model of the environment such that before taking an action, the agent can perform simulation in their ``mind". As an initial step, we assume that the agent has access to a perfect model of the environment. This allows the agent to explore the possible actions by doing beam search and back-tracking. Our results suggest that a critical piece of effective geometric planning in the real world could be learning a model of the environment and using the model for search. In addition, we do an ablation study to quantify the importance of geometrical understanding in PackIt. Our results indicate that the geometrical understanding of shapes (irrespective of planning) can significantly improve performance on PackIt. This objectively verifies that PackIt tests geometric understanding of shapes.

To summarize, our contributions are three-fold: first, we create PackIt, a virtual environment to evaluate geometric planning; second, we propose a novel way to generate challenging packing problems; and third, we design various baselines on the proposed task, including model-free learning-based and heuristic-based methods, as well as search-based black-box optimization methods.

\begin{figure}[t]
  %\centering
  \includegraphics[width=\columnwidth]{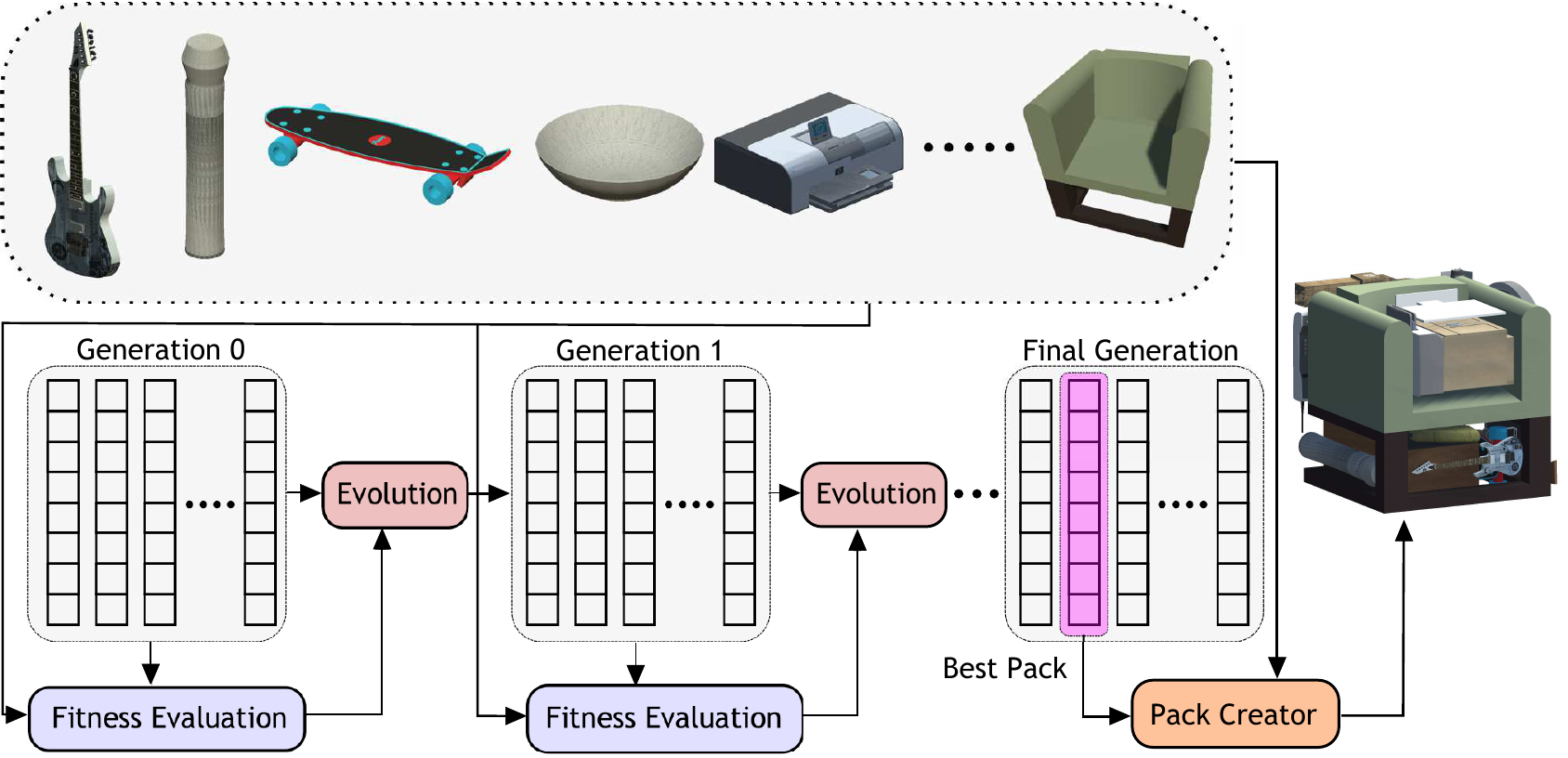}
  \caption{A schematic representation of the proposed evolutionary algorithm for generating packs. Given a pool of shapes as input, it selects a subset of shapes and appropriately scales, rotates and places them inside a unit box in order to generate high-density packs.}
  \centering
  \label{evolution}
\end{figure}
\section{Related Work}
\noindent \textbf{3D Interactive Environments in AI:} Recently, many 3D interactive environments ~\cite{Kempka2016ViZDoom,savva2017minos,wu2018building,krishnaswamy2019combining,pustejovsky2019situational,xia2018gibson} have been proposed in the AI literature. However, a majority of them have centered around the navigation task. Undoubtedly, navigation is a challenging problem in its own right, however, unlike our task, it does not require reasoning about geometrical intricacies of objects or manipulating them.

Some interactive environments like AI2-THOR~\cite{ai2thor}, HoME~\cite{brodeur2017home} and CHALET~\cite{yan2018chalet} also involve object interaction alongside visual navigation. However, in these environments, the geometrical interactions with objects are coarse, and mostly it is sufficient to treat all objects like blocks. In other words, the geometrical variability of objects does not play much role in the interaction.  Our work, on the other hand, focuses on finer geometrical interaction with the objects (like at what orientation would two complementary shapes fit into each other). Hence, our task requires more fine-grained geometrical understanding and planning. 

Simulated environments in robotics~\cite{alomari2017natural,alomari2017learning,plappert2018multi,fan2018surreal} also support interacting with objects. However, in them, the focus is on the low-level continuous control problem, and hence the interaction is generally limited to simple shapes. In contrast, an agent in PackIt interacts with and reasons about several real-world shapes simultaneously. We also choose to abstract away the low-level control problem with higher-level actions so that we can evaluate geometric planning without getting side-tracked with difficulties in orthogonal problems like grasping and kinematics. Doing packing while taking care of low-level continuous control would be an exciting direction for future work.

\noindent \textbf{Learning for Combinatorial Optimization:} There has been a recent push towards using deep reinforcement learning and imitation learning to solve combinatorial optimization problems~\cite{vinyals2015pointer,bello2016neural,bengio2018machine,gu2018pointer} like the Travelling Salesman Problem (TSP) and Knapsack. Our packing task can also be considered a combinatorial optimization problem (as we discretize the space of possible rotations and locations for placing an object inside a box). However, it differs from prior combinatorial optimization problems tackled with learning in two aspects. First, it is grounded in real-world objects and deals with skills used by humans in everyday life. Second, in addition to the optimization aspect, it also requires reasoning about the geometrical properties of real-world objects.

\noindent \textbf{Packing without learning:} There has been work done in the engineering design literature that uses genetic algorithms to search a solution for luggage packing problems ~\cite{tiwari2010fast,joung2014intelligent,fadel2015packing}. In fact, the evolutionary algorithm we design for generating the dataset of packing problems is inspired by the evolutionary algorithm for packing by~\cite{tiwari2010fast}. However, solving the packing problem using genetic algorithms would be much slower as compared to solving it by learning. By our learning-based model, the solution can be found in the same order of time as it would require to evaluate the fitness of one sample (chromosome) in a genetic algorithm (see Sec.~\ref{sec:environment}). 

Packing tasks in 3D printing are also related to ours, except cases when they have additional constraints like objects should have some distance between them~\cite{araujo2019analysis}. Works in 3D printing have used genetic algorithms with heuristics (BLBF, Largest-First) similar to our baselines~\cite{gogate2008intelligent,canellidis2010effective}. However, most of them evaluated on very few packing instances~\cite{araujo2019analysis}. ~\citep{araujo2019analysis} introduced A2018, an extensive set of packing problems for 3d printing. Our works differs from them in three ways: First, in A2018, the focus is 3D printing, so it majorly has machinery parts like gears, screws; while in PackIt the focus is geometric planning in everyday tasks, so it has objects like cups, chairs. Very few objects (like bottles) are shared. Second, in A2018, it is unclear how challenging the packing problems are as performance is reported on 13 problems out of a total of 2343, whereas we ensure the tasks are challenging by using evolution. Third, in A2018, learning is not used for packing, whereas we provide an OpenAI gym environment to facilitate research on learning-based methods. We show that learning can be a viable way to do packing.

\noindent \textbf{Packing with learning:} Some recent works~\cite{hu2017solving,laterre2018ranked} use learning for 3D bin packing~\cite{martello2000three}, where the task is to minimize the surface area of the box required to pack orthogonally placed cuboids. We, on the other hand, do not restrict to cuboids but work with arbitrarily shaped 3D objects. Hence, unlike the 3D bin packing, which is more in the spirit of combinatorial optimization problems like TSP and Knapsack, solving our task requires reasoning and understanding of geometrical nuances of real-world shapes. Additionally, we propose PackIt, consisting of a dataset of packing problems and an interactive environment for packing.

\section{Dataset}
\label{sec:dataset}
In order to pose the packing task, we first create a dataset of \textit{packs}. A \textit{pack} is defined as a set of shapes with their arrangement (rotation and location) such that they would fit inside a given box. The box is of unit size \footnote{ \label{box} For the packing task, having a unit box is as general as having any arbitrary cuboidal box ($l \neq b \neq h$). This is because, if we assume the availability of 3D models, we can first scale the cuboidal box and objects, so as to get a unit box and suitably scaled objects. After solving the packing task, we can recover the solution to the original problem by rescaling.} (i.e. $l=b=h=1$) both while generating packs and later in the virtual environment.

\begin{figure}[t]
  \centering 
  \includegraphics[width=0.45\textwidth,trim={0 0 0
0.5cm},clip]{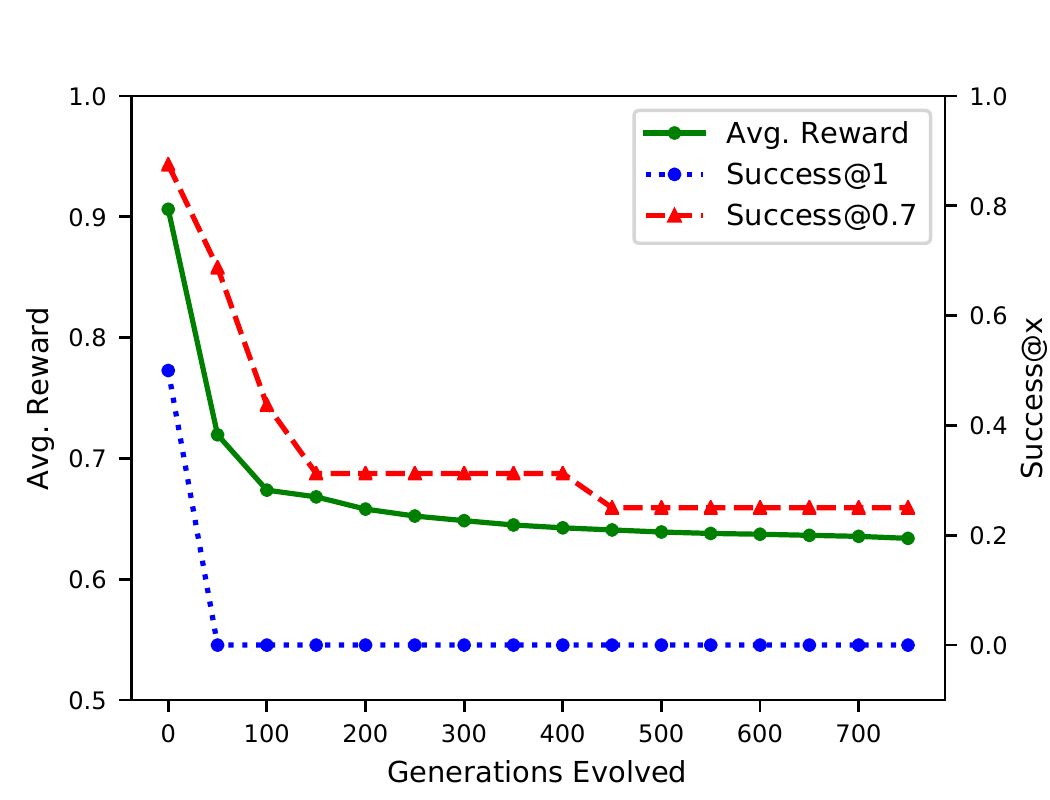}
  \label{fig:evol_acc}
  \caption{Performance of the best heuristic model on packs that are evolved to a varying extent. The left-hand y-axis is for Avg. Reward and the right-hand y-axis id for the Success@x metrics. The performance of the best heuristic decreases with evolution showing that the packs become more challenging with evolution.}
  \label{fig:evol}
\end{figure}

A dataset of packs is important because it is by deconstructing a pack (i.e. moving the shapes outside the box and rotating them back to their canonical orientation) that we generate a packing task. The existence of a pack ensures that the corresponding packing task is solvable. Apart from feasibility, we want the packing tasks to be challenging so that solving them requires geometric reasoning. For this, we design an evolutionary algorithm that generates \textit{challenging packs}. We use the density of a pack, i.e. ratio of occupied space to total space as a proxy to measure its challenging nature. We empirically show (Fig.~\ref{fig:evol}) that dense packs are hard-to-solve.

Our algorithm for generating challenging packs is inspired by the evolutionary algorithm for packing by~\citet{tiwari2010fast}, however, there are two major differences.  First, unlike~\cite{tiwari2010fast}, we don't use evolution for packing as it is much more computationally expensive than learning (Sec.~\ref{sec:environment}). We only use evolution for generating the dataset of packs, as this needs to be done only once. Second, since we are generating challenging packs, we extend the prior algorithm to support scaling objects (explained later).

Our proposed evolutionary algorithm takes a pool of shapes as input and uses them to generate a challenging pack. The algorithm is free to choose any subset of them, as well as their scale. The output is a subset of shapes from the pool, along with their size, location, and rotation that gives a dense pack. 

\begin{algorithm}[t]
    \caption{Pack Creator}
    \label{alg:pack}
    \begin{algorithmic}
        \STATE {\bfseries Input:} Box $B$, Shapes $S_1, \dots, S_n$, Chromosome $[o_1, \dots, o_n, s_1, \dots, s_n, r_1, \dots, r_n]$, 
        \FOR{$i=1$ {\bfseries to} $n$}
        \STATE Scale $S_{o_i}$ to $s_i$
        \STATE Rotate $S_{o_i}$ by $r_i$ 
        \IF{$S_{o_i}$ can be put in $B$}
            \STATE $l_i$ = best location for putting $S_{o_i}$ in $B$ according to BLBF Heuristic
            \STATE Put $S_{o_i}$ at $l_i$
       \ELSE
            \STATE $S_{o_i}$ is left outside $B$
       \ENDIF
       \ENDFOR
    \end{algorithmic}
\end{algorithm}
\noindent \textbf{Shapes: }The shapes for generating packs comes from ShapeNet~\cite{shapenet2015}, a large-scale dataset of 3D models. ShapeNet consists of $\sim 51,300$ shapes coming from 55 different categories. Since the number of possible packs is exponential in the number of shapes, we do not use the entire ShapeNet. Instead, we create three smaller subsets, one each for train, test and validation sets. These three subsets are mutually exclusive. For each object category in ShapeNet, we sample 100 shapes for train, test and validation sets. For categories with less than 300 shapes, we divided the shapes equally between the three sets. In total, for generating the packs for training, testing and validation, we have sets of 4207, 4196 and 4185 shapes respectively.

\noindent \textbf{Chromosome and Fitness Evaluator: }All the information about a pack which can be created using the given pool of shapes, is encoded in a chromosome. When we pass a chromosome and the pool of shapes through the Pack Creator algorithm (Algo.~\ref{alg:pack}) to recover the pack.  

Specifically, the \textit{chromosome} for any pack that can be created from the pool of shapes $S_1 \dots S_n$, is represented as follows:
$[o_1, \dots, o_n; s_1, \dots, s_n; \ r_1, \dots, r_n]$, where, $o_i$, $s_i$ and $r_i$ represent the order, scale and rotation for shape $S_i$. $[o_1, \dots , o_n] \in \mathfrak{S}_n$, where $\mathfrak{S}_n$ represents the set of all permutations of $\{1, 2, \dots n\}$. Also, $r_i \in \mathcal{R}$, where $\mathcal{R}$ is the set of allowed rotations. Similar to~\cite{tiwari2010fast} we allow for 24 rotations for the objects (i.e. $|\mathcal{R}| = 24$). The rotations can be derived as follows: visualize a cube with each face pained in a different color. Any color facing down results in a unique orientation. For each color facing down, there are 4 unique rotations with different colors facing front. So, in total there are $4 \times 6$ unique rotation. Further, $s_i \in \mathcal{S} =[\sfrac{1}{4}, \sfrac{1}{2}, 1, 2, 4]$. $s_i$ is defined as the ratio of the volume of the scaled object to the volume of the canonical object in ShapeNet. To sum up, for a pool of $n$ shapes, the number of possible chromosomes is $n! \times 24^n \times 5^n$.

For a given chromosome, we run the Pack Creator algorithm (Algo.\ref{alg:pack}) to determine the shapes that end up in the pack, as well as their location and rotation. Essentially, this algorithm selects shapes from the pool one at a time, in the order encoded by the chromosome. It rotates and scales the selected shape according to the chromosome. It then looks for the location in the box for this shape, described by the lowest height-width-depth (preference given to height, width, depth in order). This is also referred to as the Bottom-Left-Back-Fill (BLBF) heuristic~\cite{tiwari2010fast}. If no location is found for placing the shape, then the shape does not end up as a part of the pack. Note that for finding a location, we define a $25 \times 25 \times 25$ grid inside the box. Once we are through with all the shapes in the pool, the pack emerges. The fitness of the chromosome is determined by the ratio of the volume occupied inside the box by the total volume of the box. 

\noindent \textbf{Evolution}: For generating a pack, we start with a pool of 50 shapes. These shapes are sampled from the corresponding subset (train/test/val) of ShapeNet we created. Instead of directly sampling shapes from the dataset, we first uniformly sample a category and then sample a shape from that category. This is done as the number of shapes per category is different while we want equal representation per category.

During evolution, we maintain a population of 100 samples per generation. 25 samples with the highest fitness advance to the next generation. From the remaining 75, 25 advance to the next generation on the basis of random chance, to ensure diversity in the population. From the 50 samples that advanced to the next generation, 50 new samples are created via crossover and mutation. For creating each new sample, we randomly choose two samples out of the 50, and cross-over their chromosomes. We use ordered crossover 
(OSX)~\cite{davis1985applying} for the orders ($[o_1 \dots o_n]$) and single-point 
crossover~\cite{holland1992adaptation} for shape rotations ($[r_1 \dots r_n]$) and scales ($[s_1
\dots s_n]$). We also add mutations to the chromosome created by crossover. Specifically, we add ordered mutations (randomly swapping orders of a pair of shapes in a chromosome) and point mutations on the rotation and scale of shapes. We run evolution for a maximum of 1000 generations, with early stopping if the fitness of the best pack does not improve in the last 100 generations.

\noindent \textbf{Implementation Details: }We use Unity~\cite{goldstone2009unity} to implement the evolution algorithm. The evolution process is computationally expensive, and for generating one pack, it takes around 10 hours for a system with 2 CPUs(Intel(R) Xeon(R) CPU E5-2680 v2 @ 2.80GHz) and 16GB RAM. Our dataset contains 991 training, 515 testing and 503 validation packs, with an average of 22.6 shapes per pack.
\section{Environment}
\label{sec:environment}
\begin{figure}
  \centering
  \includegraphics[width=\columnwidth]{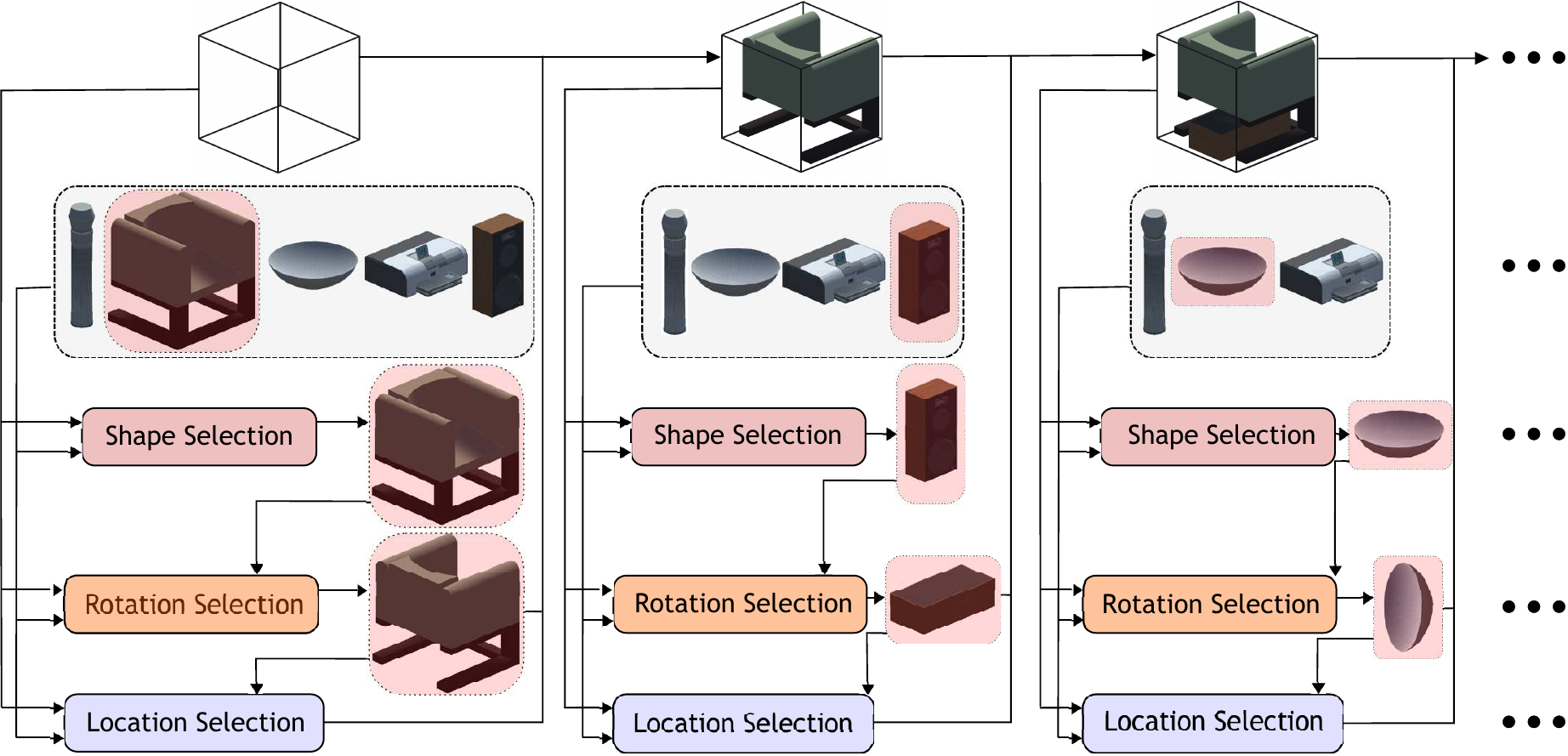}
  \caption{A schematic representation of the PackIt environment. Given a pool of shapes and a box, an agent needs to first select a shape, then rotate it and finally place it in the box.}
  \centering
  \label{fig:env}
\end{figure}
We build an interactive environment for the packing task. In our environment, packing is split into three sub-tasks, namely \textit{Shape-Selection}, \textit{Rotation-Selection}, and \textit{Location-Selection} (refer to Fig.~\ref{fig:env}). The three sub-tasks differ in their observation space as well as action space. Given a pool of shapes and a box, an agent needs to first select a shape (Shape-Selection), then rotate it (Rotation-Selection) and finally place it in the box (Location-Selection). This sequence continues until a terminal condition is met. For the feasibility of solutions, the set of rotations and locations for the agent to choose from are same as those used while generating the dataset.

\noindent \textbf{Shape-Selection Step:} During this step, an agent needs to choose an object from the ones not yet placed in the box. The agent observes the 3D representation of the box (including the objects inside it) as well as all the objects outside the box.  

\noindent \textbf{Rotation-Selection Step:} In this step, the agent has to decide a rotation for the chosen shape. The environment provides the voxel representations of the chosen shape, the box (with shapes inside it) and the shapes outside.

\noindent\textbf{Location-Selection Step:} Here the agent needs to decide where to put the shape it has already chosen. The environment provides the agent with plausible locations inside the box where the chosen shape, in the chosen rotation, can be put. The agent needs to choose one location out of the possible ones. In addition to the plausible locations, the agent also observes the voxel representations of the chosen shape, the box and the objects outside the box. 

Note that we chose to provide the agent with plausible locations for placing the object, instead of asking the agent to figure it out, as this can be computed algorithmically. We want the agent to focus on abilities like shape understanding, and not on rediscovering algorithm for finding shape intersections. Also, note that one might constrain the environment only provide those locations that have a free path to the top of the box to provably ensure every solution is kinematically instantiable.

\noindent\textbf{Reward:} The agent receives a reward when it successfully places an object inside the box. The reward is the ratio of the volume of the latest object placed inside to the total volume of all the objects in the sample. Hence, the maximum cumulative reward an agent can receive is 1, and this happens when the agent puts all the objects in the box.

\noindent\textbf{Terminal Condition:} An episode ends when there is no possible location inside the box, where the chosen shape can be put in the chosen rotation.

\noindent\textbf{PackIt-Easy:} PackIt requires both planning and geometric reasoning. To study how important geometric reasoning is in PackIt (regardless of the planning aspect), we create an easier variant of PackIt called \textit{PackIt-Easy}. Conceptually, in PackIt-Easy, we reduce the agent's effort for geometric reasoning. Hence, an agent's performance difference between PackIt-Easy and PackIt indicates the importance of geometric reasoning in PackIt.

Concretely, in PackIt-Easy, the order of Location-Selection and Rotation-Selection is swapped. Since the Location-Selection happens before Rotation-Selection, the environment provides the agent with all possible locations inside the box where the shape could be placed in any rotation. Therefore, in the worst case, Location-Selection in PackIt-Easy can take $O(|R|)$ times more computations as compared to PackIt (where $R$ is set of all possible rotations). Further, during Rotation-Selection in PackIt-Easy, the environment provides the agent all possible rotations in the chosen location, from which the agent chooses one. The brute force checking of possible locations in PackIt-Easy makes the task of the agent easier. This is because the agent could potentially place the object at the ``lowest" possible location without reasoning about its rotation. While for placing the object at the ``lowest" possible position in the vanilla version, the agent would have to first figure out which rotation would lead to the eventual placement of the object at the ``lowest" position. We further explain how this reduces the effort in geometric reasoning in Sect. \ref{sec:experiments}. 

\noindent\textbf{Performance Metric: }In order to measure performance on PackIt, we establish the \textit{Average Reward} and \textit{Success@x} metrics. Average Reward is the average cumulative reward across all samples of a dataset. Success@x is meant to reveal finer details about the performance. Success@x (x $\in [0, 1]$) is the percentage of tasks in a dataset for which the agent gets a cumulative reward that is greater than or equal to x. For example, Success@1 reveals the percentage of tasks that the agent completed successfully i.e. puts \textit{all} the objects inside the box.

\noindent\textbf{Implementation Details: }We build our interactive environment in Unity~\cite{goldstone2009unity} using the Unity ML-agents~\cite{juliani2018unity}. Our environment is OpenAI gym compatible~\cite{openai}, which facilitates easier adaptability. The voxel observations provided by the environment encode the relative size of objects by using voxels of same dimensions ($\sfrac{1}{100} \times \sfrac{1}{100} \times \sfrac{1}{100}$) for all of them. An object's voxel observation is then zero-padded on all sides to make a $100 \times 100 \times 100$ voxel volume (which corresponds to the size of the unit box). 

\noindent\textbf{Time taken in Learning vs Evolution for Packing: }Now that we have the environment, let's compare the time taken for packing via learning and via evolution. The computations done by the environment in our packing task are same as those for evaluating a single chromosome in a genetic algorithm (both involve going over shapes one at a time and finding a plausible location inside the box). Further, for a fixed problem size, decision-making in a typical learning algorithm is constant (certainly true for neural networks that we use). Hence, the time taken to solve one pack via learning is in the same order as the time to evaluate the fitness of one chromosome in the evolution counterpart.

\section{Model}
\label{sec:model}
We develop model-free heuristic-based as well as learning-based models for packing. Our models are composed of three policies, each for a sub-task.

\subsection{Heuristic-based Models:}
Heuristic-based models give a measure of the difficulty of the task and help to judge the performance of learning-based models. These models use heuristic-based policies for the three sub-tasks.

\noindent\textbf{Bottom-Left-Back Fill (BLBF) for Location Selection: }This heuristic was proposed by~\cite{tiwari2010fast}. Out of all the possible locations, this heuristic chooses the one at the bottom-most, leftmost and back-most location, with a preference for bottom over left over back.

\noindent\textbf{Largest-First for Shape Selection: }In this heuristic, we select the largest object (in terms of total volume) from the available ones. This heuristic is inspired by a similar heuristic for packing 2D shapes~\cite{chen2003two}. The intuition behind it is that the earlier packed larger objects leave between them spaces that are utilized by the smaller shapes later. Also, empirically this heuristic performs well when compared to a random shape selection policy (Tab.~\ref{tab:packit}).

\noindent\textbf{Aligned Shapes for Rotation Selection: }This heuristic aligns an object such that its smallest dimension is along the top-down direction, second smallest dimension is along the left-right direction and largest dimension is along the front-back dimension. Intuitively, it complements the BLBF heuristic for location selection as it ``tries'' that there is a minimum increase in the height of the overall pack on the addition of the new shape. The effectiveness of this heuristic is also supported by empirical evidence (see Tab.~\ref{tab:packit}).
\begin{table*}[t]
    \centering
    \begin{center}
    \begin{tabular}{lll|c||c|cccccc}
    \hline
    Shape  & Rotation  & Location & & Average & \multicolumn{6}{c}{Success@}  \\
     Selection & Selection & Selection & Learning & Reward & 0.5 & 0.6 & 0.7 & 0.8 & 0.9 & 1\\
     \hline \hline
     Random        & Aligned & BLBF & $\times$   & 0.419 & 33.39 & 13.59 & 3.11  & 0.19 & 0.00 & 0.00 \\
     Largest-First & Random  & BLBF & $\times$   & 0.473 & 44.46 & 20.0  & 4.66  & 0.58 & 0.00 & 0.00 \\
      Largest-First & Aligned & Random & $\times$ & 0.494 & 50.10 & 24.08 & 8.16  & 1.17 & 0.00 & 0.00 \\
     Largest-First & Aligned & BLBF & $\times$   & 0.592 & 75.15 & 49.71 & 22.33 & 5.63 & \textbf{0.58} & 0.00 \\
     PackNN        & Aligned & BLBF & $\surd$    & \textbf{0.649}  & \textbf{86.41}  & \textbf{70.01}  & \textbf{40.19}  & \textbf{8.15} & 0.00 & 0.00 \\
    \hline
    \end{tabular}
    \end{center}
    \caption{Performance of various model-free baselines on the PackIt environment. Leaning-based model outperforms a heuristic-based one indicating learning could be a viable option to acquire geometric planning skills for packing.}
    \label{tab:packit}
\end{table*}

\begin{table}[t]
    \centering
    \begin{center}
    \begin{tabular}{l||c|c|ccc}
    \hline
          & $\#$ of & Average & \multicolumn{2}{c}{Success@}  \\
    Model  & back-tracks & Reward  & 0.7 & 0.9 \\
     \hline \hline
    Heuristic & 0 & 0.592 & 22.33 & 0.58 \\
    Heuristic & 2 & 0.658 & 36.89 & 1.17 \\
    Heuristic & 4 & 0.693 & 49.32 & 1.55 \\
    Heuristic & 8 & 0.729 & 63.30 & 2.33 \\
    \hline
    Learning  & 0 & 0.649 & 40.19 & 0.00 \\
    Learning  & 2 & 0.680 & 49.32 & 0.19 \\
    Learning  & 4 & 0.695 & 54.56 & 0.19 \\
    Learning  & 8 & 0.715 & 62.33 & 0.78 \\
    \hline
    \end{tabular}
    \end{center}
    \caption{Performance of baselines when allowing access to a perfect model of the environment to backtrack actions{\footnotemark[3]}.}
    \label{tab:backtrack}
\end{table}
\subsection{Learning-based Model:}
We build a learning-based model for PackIt. In this work, we use learning only for Shape-Selection. For Rotation-Selection and Location-Selection, we use the Aligned and BLBF heuristic respectively. Our learning-based model should be treated as a proof-of-concept to show that learning can help in packing. A model where all the policies are learned is out of the scope of this paper and a promising direction for future work. 

\noindent\textbf{PackNN for Shape Selection:} We design PackNN, a deep neural network for Shape Selection. The input to PackNN is the voxel representation of the candidate shapes and the box containing the already placed shapes. The output is a probability distribution for selecting the next shape to be placed inside the box. (A schematic representation of PackNN can be found in the supplementary material.) We first extract features from the voxels of each candidate shape and the box with already placed shapes. These features (voxel features) are essentially average voxel occupancy for small cubes placed uniformly throughout the original $100 \times 100 \times 100$ voxel representation. We feed the voxel features to a series of three residual blocks~\cite{he2016deep} of fully connected (FC) layers to generate useful geometrical features. The parameters of these residual blocks are shared across candidate shapes while being different from those used for the box. 

Let the output feature of the series of FC residual blocks for the box be $\fea{b}$, and for the candidate shapes be $\sfea{s}{1}, \sfea{s}{2}, \dots, \sfea{s}{p}$ (assuming we have $p$ candidate shapes). We max-pool the features from the candidate shapes to get  $\bfea{s}$ that contains information about all the candidate shapes. We then create a holistic feature $\sfea{\bar{s}}{i}$ for each candidate shape, where $\sfea{\bar{s}}{i} = [\sfea{\bar{s}}{i}; \bfea{s}; \fea{b}] \ \forall \ i \in \{1, 2, \dots , p\}$. These holistic features are then fed through series of four FC residual blocks (parameters shared across candidate shapes) to generate $a_i$. Finally, the probability distribution for selecting the next shape is computed by $\text{Softmax}(a_1, a_2, \dots, a_p)$. A schematic representation of PackNN could be found in the Appendix Sec. A.

\noindent\textbf{Training Details:} We train PackNN using Proximal Policy Optimization (PPO)~\cite{schulman2017proximal}, a policy gradient method that achieves promising results on several RL benchmarks. We choose PPO because it has been shown to be sample efficient as well as stable for training.

For reduced-variance advantage estimation, we train a value prediction network which takes as input the voxel representations of the shapes and box to produce features $\bfea{s_v}$ and $\fea{b_v}$, similar to $\bfea{s}$ and $\fea{b}$ respectively produced by PackNN (network parameters are not shared between PackNN and value prediction network). The value prediction network then passes $[\bfea{s_v}; \fea{b_v}]$ through a series of four residual blocks to produce the scalar value estimate. We train PackNN for 200k Shape Selection steps (which took $\sim$6 days using 1 GPU, 8 parallel environments). After every 5k intervals, we evaluate performance on validation samples. Note that we used a subset of validation packs (80 packs) for faster training. We use the best validation model for testing. 

\section{Experiments and Results}
\label{sec:experiments}
\noindent\textbf{Evolution for Generating Challenging Pack:} In this experiment we show how our evolutionary algorithm leads to an increasingly hard packing task. To measure hardness, we use the performance metric of the best performing heuristic-based model (i.e. Largest-First for Shape Selection, Aligned for Rotation Selection and BLBF for Location Selection) (refer to Tab.~\ref{tab:packit}) as a proxy. Specifically, we start 16 different instances of pack evolution, and for each instance, we store the best pack generated at intervals of 50 generations. The best packs at different generations are meant to represent the extent to which evolution has progressed. Finally, we end up with 16 packs for each \nth{50} generation ($0, 50, 100, \dots, 750)$. In Fig.~\ref{fig:evol}, we plot Average Reward, Success@1, and Success@0.7 for the best-performing heuristic model on these packs vs the generation they correspond to. We observe that as the packs evolve, the performance of the heuristic model becomes worse. This shows how packing tasks become increasingly challenging with evolution.

\begin{figure*}
    \centering
    \includegraphics{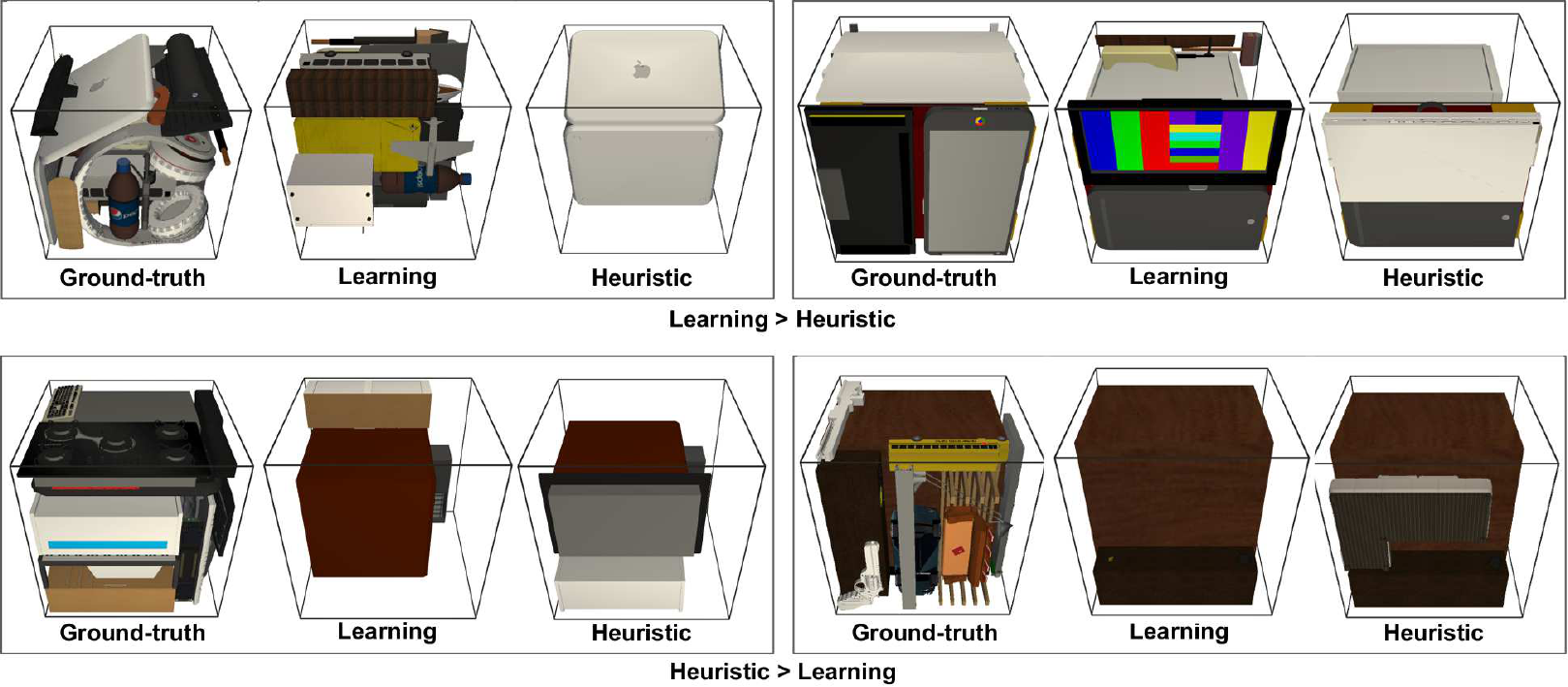}
    \caption{Some visualizations of packs in PackIt. Groud-truth shows one packing solution. The first row shows cases where Learning\footnotemark[3] outperforms Heuristic, while the second row shows vice-versa.}
    \label{fig:visual}
\end{figure*}

\noindent\textbf{Performance of Model-free Learning-based and Heuristic-based Baselines on PackIt:} Tab.~\ref{tab:packit} summarizes the performance of various model-free baselines on the test set of the PackIt dataset. Results show that our learning-based model outperforms the heuristic-based models. This is promising as it indicates that learning can be a viable option to acquire packing skills. Fig.~\ref{fig:visual} shows some qualitative results for success and failure cases of the learning-based model when compared to heuristic-based one.

To get a better idea of how much is learned via training, we can look at the performance difference between the learning-based model and the corresponding heuristic-based model with a random Shape-Selection policy. This is a fair check because when training starts, the Shape-Selection mechanism is random owing to the random initialization of parameters. We observe more than 50\% relative increase in Average Reward after training, indicating that the training process is able to discover useful skills for packing.

The results also verify the effectiveness of the Largest-First, Aligned and BLBF heuristics. When compared to a Random policy, with everything else being equal, they lead to a significant improvement in performance. 

\noindent\textbf{Performance of Baselines on augmenting with a Model of the Environment:} We explore augmenting the baselines with a model of the environment so that an agent could simulate the possibilities in its ``mind" before executing them. As an initial step, we allow the agent to have access to a perfect model of the environment.  This allows the agent to do beam search (i.e. use multiple instances of the environment) and backtrack actions (i.e. revert actions upon looking at the reward from the environment). For both the learning and heuristic baselines, we do beam search and backtracking on the shape selection. We summarize the performance of the baselines for different beam sizes and back-track budgets in Tab.~\ref{tab:beam} and Tab.~\ref{tab:backtrack}, where \textit{Learning} refers to PackNN-Aligned-BLBF model and \textit{Heuristic} refers to Largest-First-Aligned-BLBF model. We restrict the beam size to 4 as the models become very slow thereafter. 

As expected, the performance of both the baselines increases upon allowing for a larger search budget (i.e. more beam size and more backtrack budget). We also observe that the performance gap between heuristic and learned models closes and after a point heuristic outperforms learning. This is expected as the learning-based model is trained in a model-free setting. In real life, we might not have access to a perfect model of the environment, and learning a model of the environment to do an effective search might be a crucial component for geometrical planning. The current learning-based baseline lacks this component.
\begin{table}[t]
    \centering
    \begin{center}
    \begin{tabular}{l||c|c|cc}
    \hline
          & $\#$ of & Average & \multicolumn{2}{c}{Success@}  \\
    Model  & beams & Reward  & 0.7 & 0.9 \\
     \hline \hline
    Heuristic & 1 & 0.592 & 22.33 & 0.58 \\
    Heuristic & 2 & 0.632 & 31.65 & 1.17 \\
    Heuristic & 4 & 0.686 & 48.74 & 1.75 \\
    \hline
    Learning  & 1 & 0.649 & 40.19 & 0.00 \\
    Learning  & 2 & 0.697 & 54.37 & 0.19 \\
    Learning  & 4 & 0.720 & 61.75 & 0.39 \\
    \hline
    \end{tabular}
    \end{center}
    \caption{Performance of baselines when allowing access to a perfect model of the environment to do beam search{\footnotemark[3]}.}
    \label{tab:beam}
\end{table}
\begin{table}[t]
    \centering
    \begin{center}
    \begin{tabular}{ll||c|cc}
    \hline
     & &  Avg. & \multicolumn{2}{c}{Success@}  \\
     Model & Environment & Reward & 0.7 & 0.9 \\
     \hline \hline
     Heuristic&PackIt      & 0.592 & 22.33 & 0.58 \\
     Heuristic&PackIt-Easy & 0.822 & 81.74 & 33.20 \\
     \hline
     Learning&PackIt       & 0.649 & 40.19 & 0.00  \\
     Learning&PackIt-Easy  & 0.804 & 87.18 & 14.56 \\
     Learning*&PackIt-Easy & 0.830 & 88.93 & 25.04 \\
    \hline
    \end{tabular}
    \end{center}
    \caption{Performance comparison of models on PackIt and PackIt-Easy. PackIt-Easy allows for easier Rotation-Selection and therefore better performance. This shows how useful geometric understanding of shapes for Rotation-Selection could be useful for PackIt{\footnotemark[3]}{\footnotemark[4]}.}
    \label{tab:packit_easy}
\end{table}

\noindent\textbf{How useful is Geometric Understanding for PackIt?} To explore this, we create another variant of the environment, PackIt-Easy. In PackIt-Easy, Location-Selection and Rotation-Selection are swapped, so during Location-Selection, the environment provides the agent with all locations where the object could be placed in any rotation. As explained in Sec.~\ref{sec:environment}, PackIt-Easy allows the agent to place the chosen object at the ``lowest" location without having to reason about its rotation.

Consider a thought experiment wherein an agent is working in the vanilla version of the environment (i.e. Rotation-Selection before Location-Selection), and it can figure out the rotation such that the object could be placed in the ``lowest" possible location in the box. Then, the performance of this agent would be same as the performance of an agent that works in the PackIt-Easy environment with the same Shape-Selection policy and chooses the ``lowest" possible location. In Tab.~\ref{tab:packit_easy}, we show that for the same Shape-Selection policy (i.e. same heuristic policy and same learned network), a ``good" Rotation-Selection policy can boost an agent's performance. By ``good" Rotation-Selection policy, we mean a policy that selects a rotation such that the chosen object could be placed at the ``lowest" possible location. Such a Rotation-Selection policy would depend only on the chosen object and the current box configuration and would rely on the geometrical understanding of shapes. This empirically demonstrates that \textit{geometrical understanding} of shapes (irrespective of planning) is useful in PackIt.

To summarize, our experiments demonstrate the following:
\vspace{-0.2cm}
\begin{itemize}
    \itemsep0em
    \item Our proposed evolution algorithm is effective in generating challenging packs.
    \item In the model-free case, PackNN outperforms the heuristic ones, suggesting that learning can be a viable option to acquire packing skills.
    \item A crucial component for effective geometric planning could be learning a model of the environment and using that model for search.
    \item Understanding the geometry of shapes is useful in PackIt.
\end{itemize}
\vspace{-0.2cm}
\footnotetext[3]{Learning means PackNN-Aligned-BLBF and Heuristic means Largest-Fist-Aligned-BLBF.} 
\footnotetext[4]{In Learning*, PackNN is trained from scratch for Shape-Selection on PackIt-Easy}
\section{Discussion}
We proposed PackIt, a benchmark and virtual environment for testing and potentially learning geometric planning. Unlike prior 3D interactive environments in AI, PackIt requires understanding the geometry of shapes. Also, unlike robotic simulators, PackIt evaluates higher-level geometrical reasoning skills by avoiding orthogonal problems of grasping and kinematics. We also showed how we can generate challenging packing tasks using evolution. As a proof-of-concept, we demonstrated how a learning-based agent, through repeated trials, can acquire skills for packing. We also analyze search-based methods by augmenting model-free baselines with a perfect model of the environment. Our results suggest that an effective way of doing geometric planning could be learning a model of the world and using it for search. We want to emphasize that the objective of PackIt is not limited to solving the packing task but to serve as a general benchmark to evaluate geometric planning. We believe PackIt will help the community to build and test components that would enable artificial agents to do effective geometric planning in the real-world.

\textbf{Acknowledgement: } This work is partially supported by the National Science Foundation under Grant No. 1734266 and Grant No. 1903222.
\clearpage

\section*{Appendix}
\subsection*{A. PackNN}
\begin{strip}
  \newline
  \newline
  \includegraphics[width=0.99\textwidth]{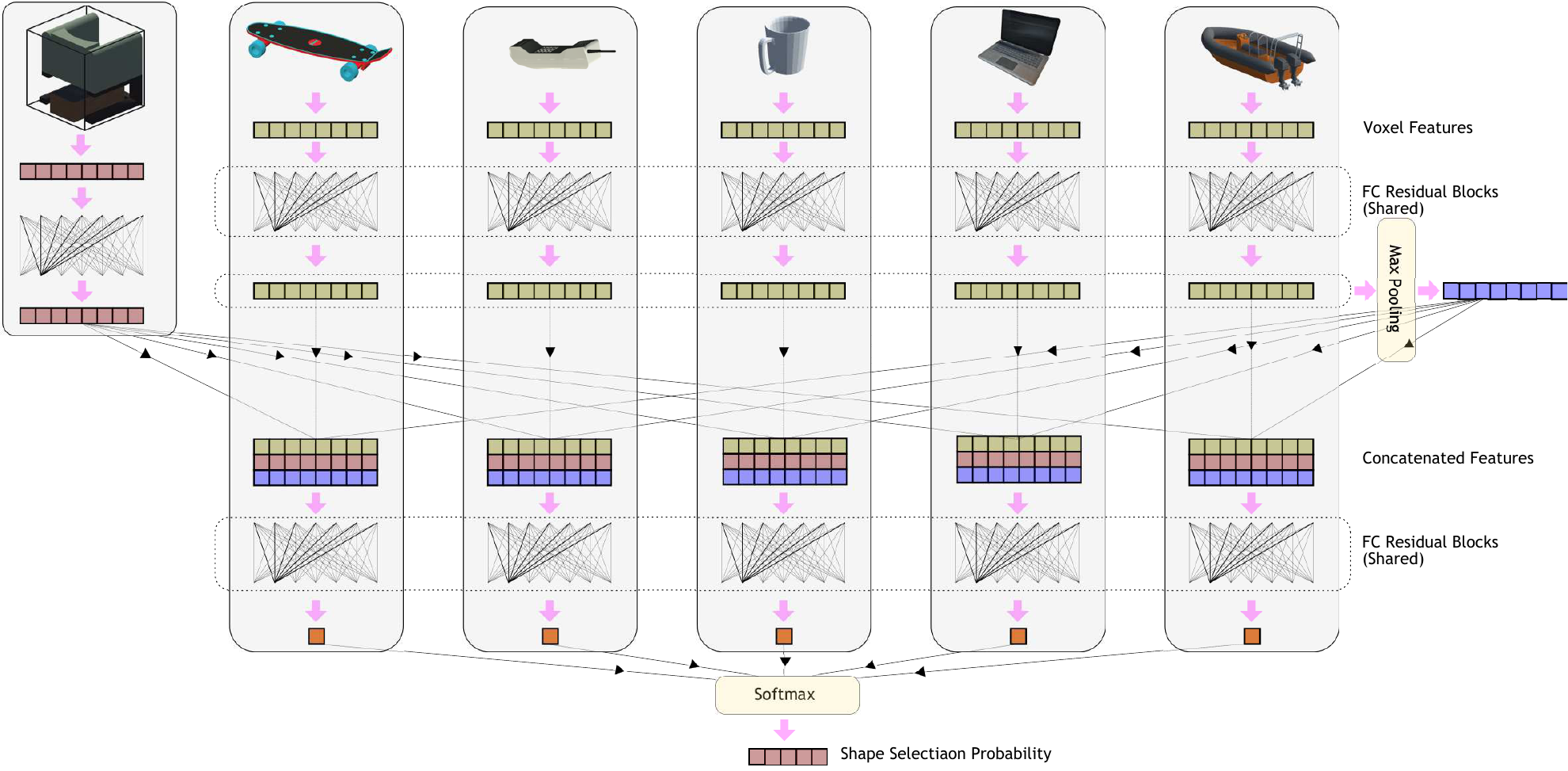}
  \captionof{figure}{PackNN model for shape selection}\label{fig:packnn}
\end{strip}
\bibliography{example_paper}
\bibliographystyle{icml2020}
\end{document}